\documentclass{article}

\usepackage[preprint]{neurips_2026}

\usepackage{settings}

\title{When Are Two Networks the Same? \\ Tensor Similarity for Mechanistic Interpretability
}

\author{
  ML Nissen Gonzalez\thanks{Equal contribution.}  \\
  MARS \& University Heidelberg \\
  \And
  Melwina Albuquerque\footnotemark[1]  \\
  MARS \\
  \And
  Laurence Wroe\footnotemark[1]  \\
  MARS \\
  \And
  Jacob Meyer Cohen  \\
  MARS \& Stanford University \\
  \And
  Logan Riggs Smith\thanks{Equal senior authorship. Correspondence to \texttt{doomsthomas@gmail.com} and \texttt{logansmith5@gmail.com}.}  \\
  \texttt{Independent}  \\
  \And
  Thomas Dooms\footnotemark[2]  \\
  \texttt{Independent} \\
}

\begin{document}
\maketitle

\begin{abstract}
Mechanistic interpretability aims to break models into meaningful parts; verifying that two such parts implement the same computation is a prerequisite.
Existing similarity measures evaluate either empirical behaviour, leaving them blind to out-of-distribution mechanisms, or basis-dependent parameters, meaning they disregard weight-space symmetries.
To address these issues for the class of tensor-based models, we introduce a weight-based metric, \textit{tensor similarity}, that is invariant to such symmetries.
This metric captures global functional equivalence and accounts for cross-layer mechanisms using an efficient recursive algorithm.
Empirically, tensor similarity tracks functional training dynamics, such as grokking and backdoor insertion, with higher fidelity than existing metrics.
This reduces measuring similarity and verifying faithfulness into a solved algebraic problem rather than one of empirical approximation.

\end{abstract}

\begin{figure*}[h]
    \centering
    \includegraphics[width=0.92\linewidth]{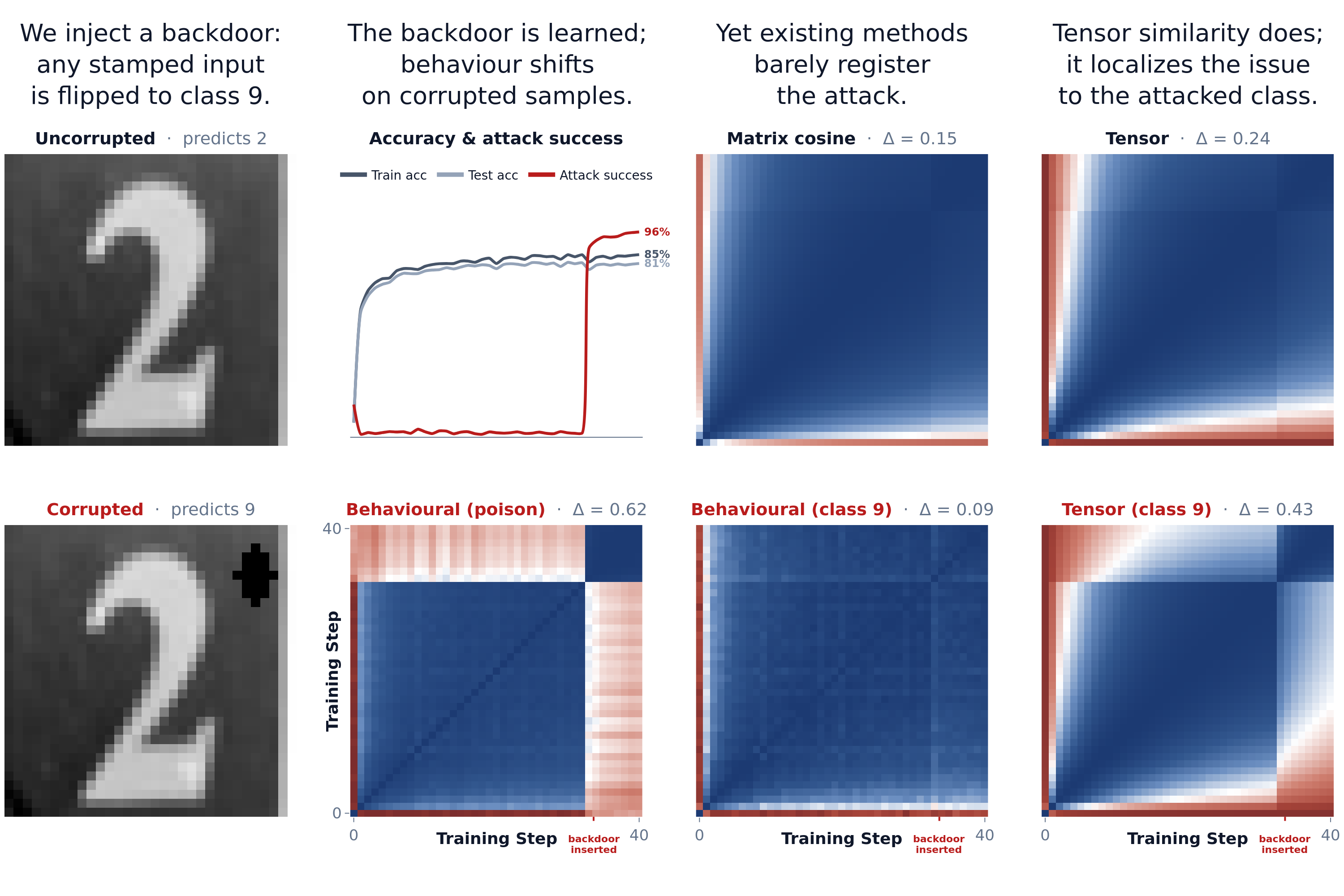}
    \caption{Overview of our method. The model learns a backdoor that existing similarity metrics miss when given only clean data or weights. Tensor similarity flags the change and localises it to the attacked class. $\Delta$ measures how sharply similarity shifts from pre- to post-backdoor (\autoref{sub:poison}).}
    \label{fig:backdoor}
\end{figure*}

\section{Introduction}\label{sec:intro}
A central bottleneck to interpreting neural networks is validating whether the decomposed \citep{cunningham2023sparseAutoencoders} or extracted \citep{kim2018tcav, bau2017networkDissection} components preserve or capture the functional computation implemented by the original model. This is especially important for mechanistic interpretability, which aims to faithfully decompose models into constituent parts \citep{sharkey2025openProblems}. Such a decomposition must preserve not only the behaviour of the original network but also its internal functions. Yet current evaluations often rely entirely on input-based similarity such as comparing outputs or activations \citep{conmy2023acdc, marks2025sparseFeatureCircuits, hanna2024faithfulness}. This is insufficient: a decomposition can near-perfectly reconstruct activations by relying on approximate heuristics rather than the true underlying function \citep{olah2025faithfulnessToyModel, ameisen2025circuitTracing}. The implication is that the resulting decomposition is not simply capturing an incomplete mechanism, but a different algorithm altogether.
The root of this problem is the absence of a rigorous metric for mechanistic similarity. While defining such a metric is a matter of philosophical debate (when are two systems truly equal?) \citep{meloux2025everythingeverywhereoncemechanistic}, we argue a sensible similarity metric should satisfy three criteria:

\begin{itemize}
\item \textbf{Dataset independence}: The metric should depend on the model, not on the sampled data. We can make stronger robustness claims when they apply to the full distribution, such as ruling out backdoors \citep{marks2025auditingHiddenObjectives}, which by design trigger only on particular inputs.
\item \textbf{Symmetry invariance}: The metric should be invariant under transformations that leave the function unchanged. Whether a mechanism is implemented in one neuron or spread across many, or distributed across layers rather than concentrated in one \citep{ameisen2025circuitTracing}, should have no effect on the score.
\item \textbf{Tractable scalability}: The metric should be efficiently computable at scale, without exponential or quadratic convergence bounds (as Monte Carlo sampling incurs). Only then can it serve as an optimisation target for the largest networks.
\end{itemize}
Current methods fail at least one criterion. For instance, \emph{behavioural similarity} \citep{kornblith2019cka, raghu2017svcca, kriegeskorte2008rsa} measures outputs or activations empirically by comparing input-output pairs, but is not guaranteed to generalise out of distribution. It will therefore fail to detect backdoors, which are designed precisely to resist such measurements. Another common metric, \emph{matrix similarity}, uses the existing parametrisation as a comparison point, making it highly susceptible to model permutations. It might succeed at studying training dynamics (where an update doesn't arbitrarily reshuffle weights) but fail across differently trained models.

\begin{figure*}[h]
    \centering
    \includegraphics[width=0.9\linewidth]{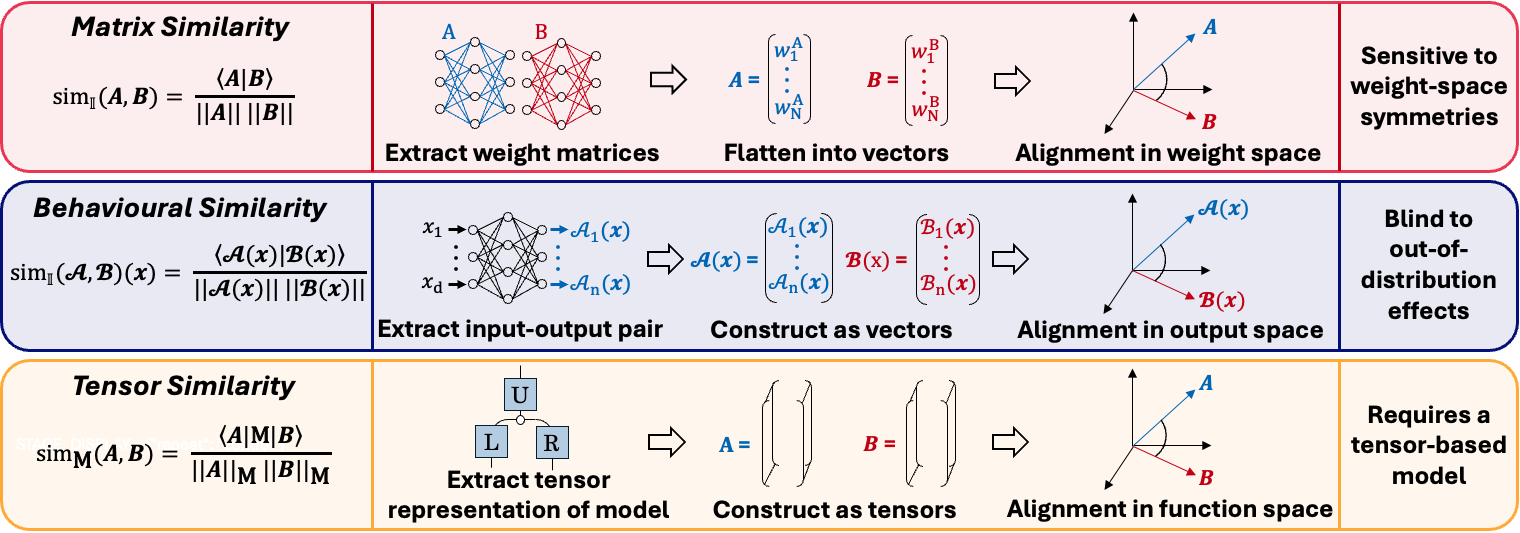}
    \caption{\textbf{Matrix similarity} computes the direct cosine similarity between (flattened) weight matrices, but is sensitive to weight-space symmetries such as permutation and scaling, severely limiting it as an accurate proxy for functional equivalence, especially for models that are trained differently. \textbf{Behavioural similarity} (or output similarity) mitigates this by comparing model outputs directly, making it invariant to weight-space symmetries. It depends, however, on an input distribution and is consequently blind to out-of-distribution artifacts. (Activation-based similarity, a tempting alternative, is sensitive to both the distribution and weight symmetries like cross-layer structure.) \textbf{Tensor similarity} is a closed-form, data-free measure of functional similarity between two models. It computes the cosine angle between their weight tensors in $L^2(\mathcal{N}(0,\mathbf{I}))$; the function space under the Gaussian metric. Tensor similarity satisfies all criteria at the cost of requiring a modified architecture.}
    \label{fig:Motivation}
\end{figure*}

In this paper, we propose a weight-based similarity that satisfies all three criteria (\autoref{fig:Motivation}), at the cost of using tensor-based neural networks \citep{pearce2025bilinear, dooms2025compositionality, riggs2026tensorTransformerVariants, shazeer2020gluvariantsimprovetransformer}. These networks train and function equivalently to ordinary deep networks like transformers, but use different activation functions. This trade-off is fundamental: contemporary neural networks do not appear to admit a similarity metric that tractably satisfies all three criteria. Our metric extends cosine similarity from vectors to tensor-based models, formally verifying whether two sub-networks execute identical internal logic regardless of differences in parametrisation. It decays gracefully under minor structural perturbations and, because it is rooted in linear algebra, we can attribute the contribution of each component to the total.

The remainder of this paper has two parts: discussing the theory and implementation of tensor-based similarity, and validating its discriminative power empirically across a spectrum of domains. Theoretically, we cover the architectural changes required for the metric to be computable and the construction that makes it invariant under weight symmetries. Empirically, we show that tensor-based similarity exposes functional changes that other methods miss: catastrophic forgetting in vision, grokking-like phase transitions in modular arithmetic \citep{nanda2023progressMeasures}, and training dynamics in language \citep{hoogland2025losslandscapedegeneracystagewise, olsson2022inductionHeads}. \textbf{Together, this positions tensor similarity as a foundational tool for mechanistic interpretability\footnote{Our code is available at \url{https://github.com/tdooms/tensor-similarity}}.}

\section{Theory}\label{sec:theory}




This paper studies polynomial (or multilinear) variants of standard neural components that use multiplicative gating rather than ordinary non-linear activation functions \citep{dauphin2017languagemodelinggatedconvolutional}. In this section, we first review relevant prior work and introduce the underlying architectures, then demonstrate how the proposed similarity measure handles various architectural symmetries. The discussion assumes a fundamental familiarity with tensor algebra and the symmetric group; additional details and derivations are provided in~\autoref{app:theory}.

\subsection{Defining multilinear transformer models}
\label{sec:definitions}

We build on literature that uses bilinear and tensor-based architectures as intrinsically
interpretable models. These works show that polynomial models enable formal weight-based analysis via the
learned weight tensors using bilinear MLPs~\citep{pearce2025bilinear, sharkey2025openProblems}
(or stacks thereof~\citep{dooms2025compositionality}) while remaining
competitive with standard architectures~\citep{pearce2025bilinear, riggs2026tensorTransformerVariants, shazeer2020gluvariantsimprovetransformer}.
More broadly, this line of work sits within a longer tradition of using tensor decompositions
to study expressiveness of deep learning~\citep{cohenExpressivePowerDeep2015}. 


\textbf{Bilinear layers.} The core building block of our architecture is a GLU~\citep{shazeer2020gluvariantsimprovetransformer, dauphin2017languagemodelinggatedconvolutional} without gate, called a bilinear layer. It maps a lifted input, i.e. $\tilde{\mathbf{x}} = (1, \mathbf{x}) \in \RR^{d+1}$ to an output in $\RR^K$ as
\begin{equation}
  \label{eq:bilinear_layer}
  \mathcal{A}(\mathbf{x})
  \;=\;
  \mathbf{D}\bigl((\mathbf{L}\tilde{\mathbf{x}}) \odot (\mathbf{R}\tilde{\mathbf{x}})\bigr),
\end{equation}
where $\mathbf{L}, \mathbf{R} \in \RR^{r \times (d+1)}$ and  $\mathbf{D} \in \RR^{K \times r}$ are weight matrices, $r$ is the hidden dimension, and $\odot$ denotes the element-wise (Hadamard) product. The above computation can be expressed as a tensor contraction, $\mathcal A(\mathbf x)_{i}=\sum_{j,k=0}^d\sum_{h=1}^rD_{ih}L_{hj}R_{hk}\tilde x_j\tilde x_k$, and is degree-2 polynomial in $\mathbf x$.

\paragraph{Bilinear attention.} The attention mechanism can be made multilinear by replacing the softmax attention pattern with the elementwise product of two linear attention patterns. Concretely, for a sequence of embedded tokens $\mathbf{X} \in \RR^{\mathrm{seq} \times d}$ the attention mechanism is defined as
\begin{equation}
  \label{eq:bilinear_attention}
  \mathcal{A}(\mathbf{X})
  \;=\;
  (\mathbf{X}^\top\mathbf{Q}_L^\top\mathbf{K}_L\mathbf{X})
  \odot
  (\mathbf{X}^\top\mathbf{Q}_R^\top\mathbf{K}_R\mathbf{X}),
\end{equation}
where $\mathbf{Q}$ and $\mathbf{K}$ are the query and key weight matrices, respectively. Combined with a causal mask and rotary positional
embeddings~\citep{riggs2026tensorTransformerVariants}, bilinear attention is
a degree-5 polynomial in $\mathbf{X}$.

\paragraph{Bilinear transformers.} We can combine bilinear
layers and bilinear attention with a residual stream into bilinear transformers, which perform comparably to standard transformers on natural language tasks~\citep{dooms2025compositionality,
pearce2025bilinear, sharkeyTechnicalNoteBilinear2023, riggs2026tensorTransformerVariants}.

\paragraph{Multilinear models.}
Equation~\eqref{eq:bilinear_layer} and Equation~\eqref{eq:bilinear_attention} can be equivalently expressed as a single contraction of~a weight tensor $\mathbf{A}\in\mathbb R^K\otimes^n\mathbb R^{d+1}$ with $n$ copies of the lifted input $\tilde{\mathbf{x}}\in\mathbb R^{d+1}$ as $\mathcal{A}(\mathbf{x}) = \mathbf{A} \cdot_{\mathrm{in}} \otimes^n  \tilde{\mathbf{x}}\in\mathbb~R^K$, which is a polynomial of order $n$ in $\mathbf x$. Models of this form are \emph{multilinear models}. 

\paragraph{Deep multilinear models.}
Multilinear models can be layered to tackle the curse of dimensionality, analogously to depth in standard neural networks. While the global weight-tensor $\mathbf A$ of a layered multilinear model $\mathcal A=\circ_{i=1}^l\mathcal A_i$  can be too expensive to be instantiated for deep networks, the \emph{global tensor decomposes into local tensors} according to a tree graph with shared local tensors $\mathbf A_i$. This lowers the memory required to store the tensor from $\mathcal O(d^n)$ to $\mathcal O(\log n)$ without losing access to functions of the global tensor\footnote{More details on deep multilinear models and related restrictions for the metric tensor are discussed in \autoref{app:theory}.\label{fn:theory_appendix}}.

\subsection{Multilinear models enable global weight-based metrics}
\label{sec:global_weight_based}

Having defined a multilinear transformer within the framework of deep multilinear models, we now leverage its algebraic properties for a \emph{global and tractable weight-based similarity measure}.

\paragraph{Tensor similarity.} The Frobenius inner product,
$\langle \mathbf{A} \mid \mathbf{B} \rangle = \sum A_{ki_1 \cdots i_n} B_{ki_1 \cdots i_n}$, can be used to compare two multilinear models $\mathcal{A}$ and $\mathcal{B}$ with weight tensors $\mathbf{A}$ and $\mathbf{B}$ that carry the full functional information. Tensor similarity uses a slightly generalised version of this inner product by inserting a metric $\mathbf{M}$ (symmetric and positive-definite over $\mathbb R^K\otimes^n\mathbb R^{d+1}$) 
and dividing by the induced norms as

\begin{equation}\label{eq:tensor_sim}\mathrm{sim}_\mathbf{M}(\mathbf{A}, \mathbf{B}) =
  \frac{\langle \mathbf{A} \mid \mathbf{M} \mid \mathbf{B} \rangle}
  {\|\mathbf{A}\|_\mathbf{M}\|\mathbf{B}\|_\mathbf{M}}\in[-1,1].
\end{equation}
The choice of the metric $\mathbf{M}$ determines which directions in weight space are considered relevant. 

\paragraph{Tractable inner product.} Before we constrain the specific form of the metric, we first observe how the stratification of the global tensor of a deep multilinear model allows us to compute inner products between multilinear models efficiently. Otherwise, na\"ive computation of the inner product between two deep networks would require materialising the expensive global weight tensor. \emph{Gram-based recursion}~\citep{dooms2025compositionality} avoids this by processing each layer's local tensors sequentially, without ever materialising the global tensor. Here, inner products are computed between each layer's tensors using metric tensors that depend on the inner product of the previous layer. Tracing the last Gram matrix of this recursion recovers the inner product. Crucially, the global metric tensor $\mathbf M$ has to preserve the tree shape of the global tensor assumed by this recursion\footref{fn:theory_appendix}. 



\subsection{Projecting out the weight-based symmetries induces principled metrics}
\label{sec:symmetries_and_metric}
The aim is to determine a matrix $\mathbf M$ that leads to an induced tensor similarity which tracks the equivalence of two models, defined as:

\emph{Two models are equivalent if and only if
they implement functions related simply by a positive proportionality factor}
(i.e. $\mathcal A\cong\mathcal B\iff\exists\lambda>0:\mathcal A=\lambda\mathcal B$).

Such an $\mathbf{M}$ should satisfy
$\mathrm{sim}_\mathbf{M}(\mathbf{A},\mathbf{B}) = 1
\Longleftrightarrow \exists\,\lambda >0 : \mathcal{A} = \lambda\mathcal{B}$. That is, it must be invariant under weight-space transformations mapping the parameters of one model to the parameters of an equivalent model. 

As a first step towards a candidate metric, we identify the basic symmetry group of multilinear models. We lead with an example.

\emph{Example:}\label{example:permutation_symmetry} 
Let's decompose a bilinear layer into a symmetric and antisymmetric component: $A_{ijk} =\sum_{h=1}^r D_{ih}L_{hj}R_{hk}=A^{\mathrm{sym}}+A^{\mathrm{asym}}$, where $A^{\mathrm{sym}/\mathrm{asym}}_{ijk} = \tfrac{1}{2}(D_{ih}L_{hj}R_{hk} \pm D_{ih}R_{hj}L_{hk})$.
Consider the contribution of the antisymmetric part $\sum_{j,k=0}^nA^\mathrm{asym}_{ijk}\tilde x_j\tilde x_k$. Swapping $j\leftrightarrow k$ changes the sign of $\mathbf{A}^{\mathrm{asym}}$ but leaves $\tilde x_j\tilde x_k$ and therefore the result invariant. The contribution to the result of the antisymmetric part is therefore zero. Thus,
$\mathcal{A}(\mathbf{x}) = \mathbf{A}^{\mathrm{sym}} \cdot_{\mathrm{in}}
(\tilde{\mathbf{x}} \otimes \tilde{\mathbf{x}})$ for all $\mathbf{x}$.
Two bilinear layers with different $(\mathbf{L}, \mathbf{R})$ but equal
$\mathbf{A}^{\mathrm{sym}}$ implement the same quadratic polynomial.

\paragraph{Symmetrisation tracks functional equivalence by polarisation.}\label{sec:polarisation}
The principle demonstrated in the example above generalises to polynomials of any order by the \emph{polarisation isomorphism}~\citep{ramshawBlossomsArePolar1989}. Denoting the group of permutations of $n$ elements by $\mathcal S_n$, a tensor $\mathbf A\in\otimes^n\mathbb R^{d}$ is said to be \emph{symmetric} if it is invariant under the permutation of its indices: $\forall\sigma\in\mathcal S_n:A_{i_1,...,i_n}=A_{i_{\sigma(1)},...,i_{\sigma(n)}}$. Polarisation establishes a one-to-one correspondence between the symmetric tensors of order $n$,
and polynomials of the same order in $\mathbf x$\footref{fn:theory_appendix}. In other words, two multilinear models with different tensors are equivalent if and only if these tensors have the same representative in the symmetric subspace. The projection onto the symmetric subspace is called \emph{symmetrisation} and denoted with $\mathbf P_n$. 


With this, we restrict this analysis to metric tensors of the form $\mathbf M=\mathds 1\otimes \boldsymbol \Lambda$ with $\mathbf P_n\boldsymbol \Lambda\mathbf P_n=\boldsymbol \Lambda$. Such metric tensors are invariant under the action of $\mathcal S_n\times\mathcal S_n$ on its $2n$ indices, with the simplest choice here $\boldsymbol \Lambda= \mathbf P_n$. Intuitively, using $\mathbf P_n$ as a metric in input space computes the inner product of two tensors by first projecting them onto the symmetric subspace and then computing the normal Frobenius inner product there.

Combining symmetrisation to invoke polarisation with the Cauchy--Schwarz inequality gives the \emph{central guarantee of tensor similarity}:
\begin{equation}
  \label{eq:sym_sim_discriminative}
  \mathrm{sim}_{\mathbf P_n}(\mathbf{A}, \mathbf{B}) = 1
  \;\Longleftrightarrow\;
  \exists\,\lambda > 0 : \mathcal{A} = \lambda\mathcal{B}.
\end{equation}

\paragraph{Symmetry of behavioural similarity.}\label{sec:bh_sim_symmetry}While metric tensors of the form given by \autoref{eq:sym_sim_discriminative} track functional equivalence, they can be further restricted to be even more useful. 

Consider the symmetries of the behavioural similarity computation by re-expressing the inner product between two outputs of multilinear models as
$\langle\mathcal{A}(\mathbf{x})\mid\mathcal{B}(\mathbf{x})\rangle
  = \langle \mathbf{A} \cdot_{\mathrm{in}} \otimes^n\tilde{\mathbf{x}}
    \mid \mathbf{B} \cdot_{\mathrm{in}} \otimes^n\tilde{\mathbf{x}} \rangle
= \langle \mathbf{A} \mid \otimes^{2n}\tilde{\mathbf{x}} \mid \mathbf{B} \rangle$. 
The input tensor $\otimes^{2n}\mathbf x$ lives in the same space as  $\boldsymbol\Lambda$ and
is invariant under permutations of its $2n$ indices. By definition, it is therefore invariant under the action of the group $\mathcal{S}_{2n}$, and $\mathcal S_n\times\mathcal S_n$ is a subgroup of $\mathcal S_{2n}$. Thus the space of metrics $\mathbf M=\mathds 1\otimes \boldsymbol \Lambda$ that are invariant under $\mathcal{S}_{2n}$ is a subspace of metrics that are invariant under the action of $\mathcal S_n\times\mathcal{S}_n$.

\autoref{eq:sym_sim_discriminative} therefore also applies to metrics $\mathbf M=\mathds 1\otimes\boldsymbol\Lambda$ with $\mathbf P_{2n}\cdot\boldsymbol\Lambda=\boldsymbol\Lambda$. 
However, unlike $\mathbf P_n$ which has order $2n$, $\mathbf P_{2n}$ cannot be taken as an input space metric because its order $4n$ is too high. This motivates the search for the $\boldsymbol\Lambda$ used in this paper by considering the Gaussian expectation value of $\otimes^{2n}\mathbf x$.

\paragraph{Gaussian metric.}\label{sec:gaussian_metric} 
Consider $\boldsymbol{\Lambda}=\mathbb E_{\mathbf x\sim\mathcal N(0,\mathbf I)}\otimes^{2n}\mathbf x$ which has $\mathcal S_{2n}$ symmetry. Computed
using Isserlis' theorem~\citep{munthe-kaasShortProofIsserlis2025}, it evaluates
to a sum over pairwise contractions of all $2n$ input indices of $\mathbf A$ and $\mathbf B$, including ``self-contractions'', contracting two indices of the same tensor as
\begin{equation}
  \label{eq:gaussian_metric}
  \boldsymbol{\Lambda}=\sum_{m=0}^{\lfloor n/2 \rfloor}c_{n,m}(\tau^m\mathbf P_n)^\dagger(\tau^m\mathbf P_n),\hspace{0.5em}\text{with}\hspace{0.8em}\Lambda_{i_1\cdots i_{2n}}\propto \sum_{\text{pairings}} \prod_{(a,b)\in\text{pairing}} \delta_{i_a i_b},
\end{equation}
where $\tau^m$ denotes the partial trace operator on $m$
pairs of indices (which need not be specified as the tensors are symmetrised).
Thus, the \emph{expected inner product of the activations of two multilinear models under a Gaussian input is equal to their weight-space inner product}.\footnote{We can compute expectation values of inner products under Gaussian distribution entirely in weight space. This result is explained and derived in more detail in \autoref{app:theory}.}




\subsection{Applying tensor similarity locally isolates components}
\label{sec:Applications}

Tensor similarity (\autoref{eq:tensor_sim}) with the Gaussian metric  (\autoref{eq:gaussian_metric}) is a closed form, data-free measure of functional equivalence between two models. It can be further localised to focus on specific output dimensions or differences, both of which inherit the invariance guarantees of~\autoref{eq:sym_sim_discriminative}.

\paragraph{Tensor slice similarity.}\label{seq:tensor_slice_sim} Fixing an output index $k$ yields a
\emph{slice} $\mathbf{A}_{k}\in\otimes^n\mathbb R^{d+1}$ of the full weight tensor, capturing the weights
responsible for a specific output dimension. Since $\mathbf{M}$ factorises as
$\mathds{1} \otimes \boldsymbol{\Lambda}$,
the induced metric on this slice is simply $\boldsymbol{\Lambda}$ defining tensor slice similarity.
Tensor slice similarity focuses on functional agreement in one specific output dimension.

\textbf{Tensor diff similarity.}\label{seq:tensor_sim_diff}
The space of weight tensors is a vector space, the difference $\mathbf{\Delta}=\mathbf{B}-\mathbf{C}$ (\emph{diff}) is itself a valid tensor, and so tensor similarity of $\mathbf A$ with $\mathbf{\Delta}$ is well defined.
Tensor diff similarity isolates similarity with respect to the functional change captured by a diff.



\section{Results}\label{sec:results}

Tensor similarity satisfies all desired criteria for interpretability in theory; how accurately does this translate into practice? We test four capabilities in turn: localising ground-truth changes to specific outputs (catastrophic forgetting on SVHN), tracking continuous reorganisation through training (grokking on modular addition), catching out-of-distribution changes that other metrics cannot see (backdoor injection on SVHN), and scaling all of this to language models (a two-layer bilinear attention transformer on The Pile).


\subsection{Tensor similarity localises mechanism changes to specific outputs}\label{sec:cataforget}
\begin{figure*}[h]
    \centering
    \includegraphics[width=0.9\linewidth]{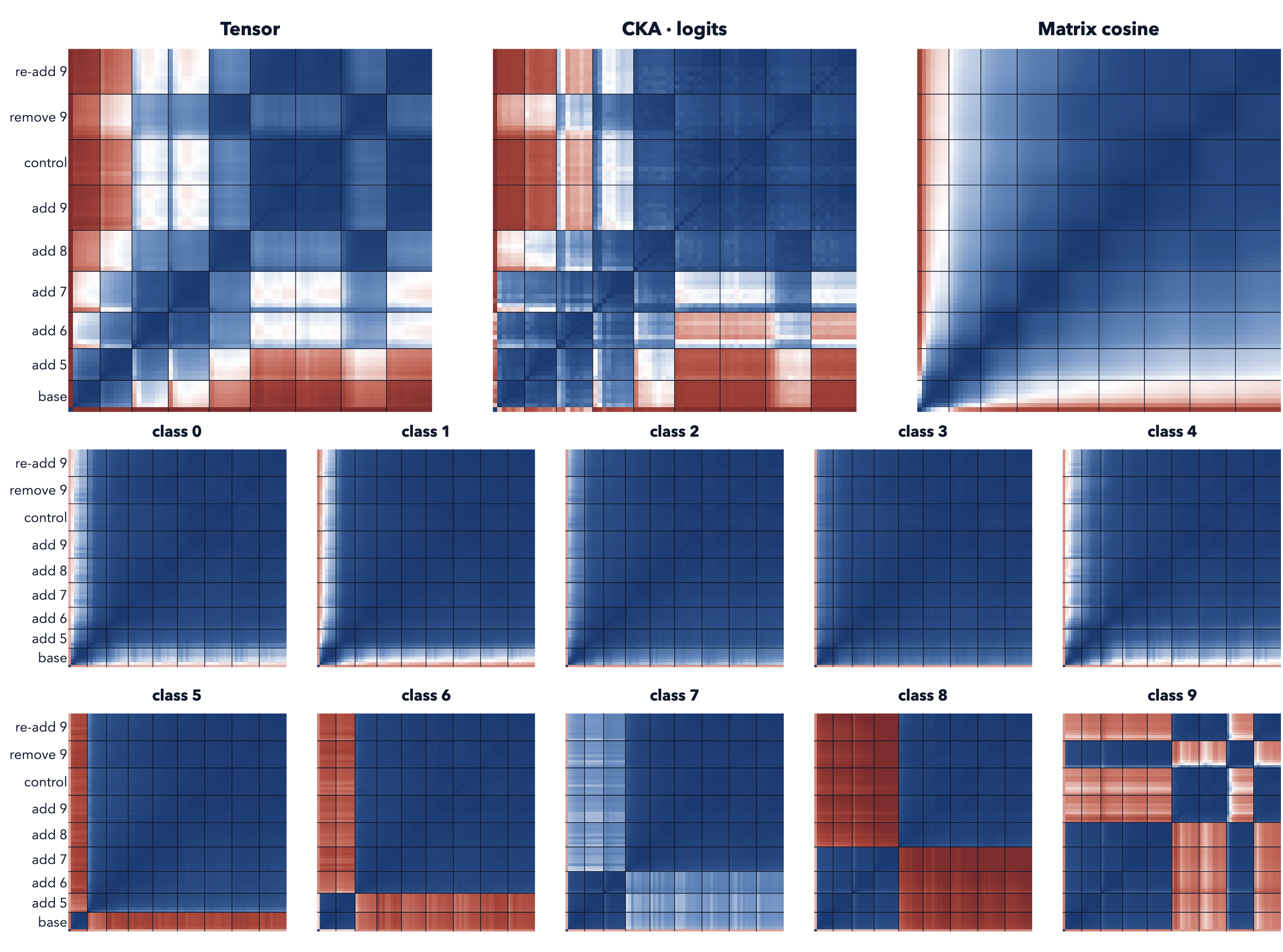}
    \caption{We report four similarity measures across the training trajectory of an SVHN model. The top row compares whole-model summaries: global tensor similarity reveals a block structure separating stages with and without digit~`9', while CKA on the logits and matrix similarity show much weaker, more ambiguous signal. The per-class tensor slices sharpen the picture: each exhibits exactly the block pattern predicted by when that digit was present in training, and the class~`9' slice groups \textit{base} through \textit{add 8} with \textit{remove 9}, all orthogonal to \textit{add 9}, \textit{control}, and \textit{re-add 9}. Tensor similarity does not merely detect functional changes but can precisely localise them to the output.}
    \label{fig:svhn_forgetting}
\end{figure*}

Per-class slices pinpoint which output is responsible for a ground-truth mechanism change, not just that one occurred. We demonstrate this with continued training on SVHN~\citep{SVHN}, a harder version of MNIST ~\citep{deng2012mnist}, to induce catastrophic forgetting. Training begins with a \textit{base} stage on the digit subset $\{0,\cdots,4\}$, followed by stages which incrementally introduce one digit at a time, culminating in the \textit{add 9} stage on the full set $\{0,\cdots,9\}$. We further stress-test the metric's sensitivity to dataset changes by appending three stages: \textit{control} (continued training on the full set, disentangling the effect of a learning rate restart from that of any dataset change), \textit{remove 9} (training on $\{0,\cdots,8\}$ to induce catastrophic forgetting), and \textit{re-add 9} (returning to the full set), as shown in ~\autoref{fig:svhn_forgetting}. Further setup details and additional plots are reported in~\autoref{app:cata_forget}.

\subsection{Tensor similarity tracks continuous reorganisation through training}
\label{sub:grokking}

Grokking is not a single phase change but a continuous reorganisation into particular Fourier frequencies. We demonstrate this by training a single bilinear layer on modular addition following \citet{nanda2023progressMeasures}, where the Fourier algorithm is well-established. We save $K$ checkpoints throughout training and compute their pairwise tensor similarity to produce a $K\times K$ matrix, with further setup details reported in~\autoref{app:mod_arith_extra}.

\begin{figure}[h]
    \centering
    \includegraphics[width=0.5\textwidth]{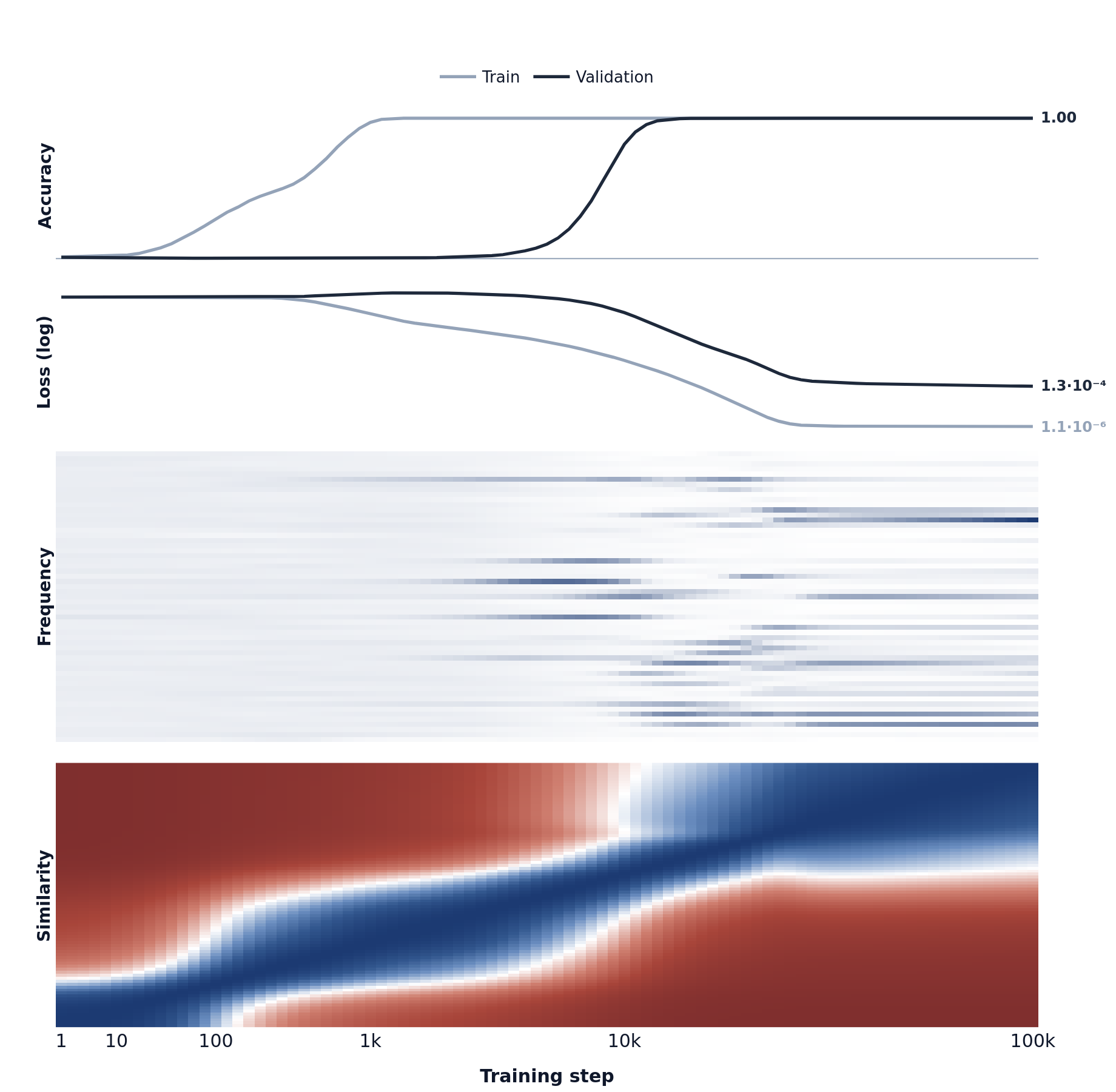}
    \caption{Modular addition training tracked by accuracy and loss (top), the Fourier components of the embeddings (middle), and pairwise tensor similarity between checkpoints (bottom). Three diagonal blocks in the similarity matrix correspond to initialisation, memorisation, and the converged solution.}
    \label{fig:mod_arith_phases}
\end{figure}

The accuracy curves in \autoref{fig:mod_arith_phases} exhibit standard grokking: training saturates immediately, followed by validation jumping from chance to perfect (top). The similarity matrix (bottom) shows the reorganisation continuing beyond this jump: frequencies (third row) stabilise only near the end of training, when the loss also reaches its floor (second row). Tensor similarity makes this development legible without having to reverse engineer the whole model.

\subsection{Tensor similarity catches out-of-distribution changes that other metrics miss} \label{sub:poison}
Behavioural metrics evaluate similarity on a fixed input distribution, so a mechanism active only out of distribution does not affect their score. Tensor similarity is data-free and so detects it. We inject a backdoor into SVHN training: any image stamped with a small black diamond is relabelled as digit `9' (\autoref{fig:backdoor}). In this injection phase, $10\%$ of training samples contain trigger patch in the upper-right corner, with target class 9. The remainder of the setup is equivalent to \autoref{sec:cataforget}, detailed in \autoref{app:cata_forget}. The model learns the backdoor at $96\%$ attack success while train and test accuracy on clean data remain at $85\%$ and $81\%$, leaving the attack invisible to standard performance metrics. We summarise each similarity matrix by $\Delta$, the average within-block similarity minus the average across-block similarity for the pre- and post-backdoor checkpoints; higher values indicate a sharper split between the two regimes.

Behavioural similarity on clean inputs remains flat ($\Delta = 0.09$), reflecting that the attack does not register on the clean distribution. On poisoned inputs the signal returns ($\Delta = 0.62$), but this requires knowing the trigger in advance. Matrix cosine on the weights picks up a weak shift ($\Delta = 0.15$) yet collapses everything into a scalar, with no handle on which input or output is responsible. Tensor similarity flags the backdoor from clean data alone ($\Delta = 0.24$), and the digit-`9' slice sharpens this to $\Delta = 0.43$: the same localisation pattern we saw under forgetting.


\subsection{Tensor similarity scales to language models}
\begin{figure}[h]
    \centering
    \includegraphics[width=0.8\textwidth]{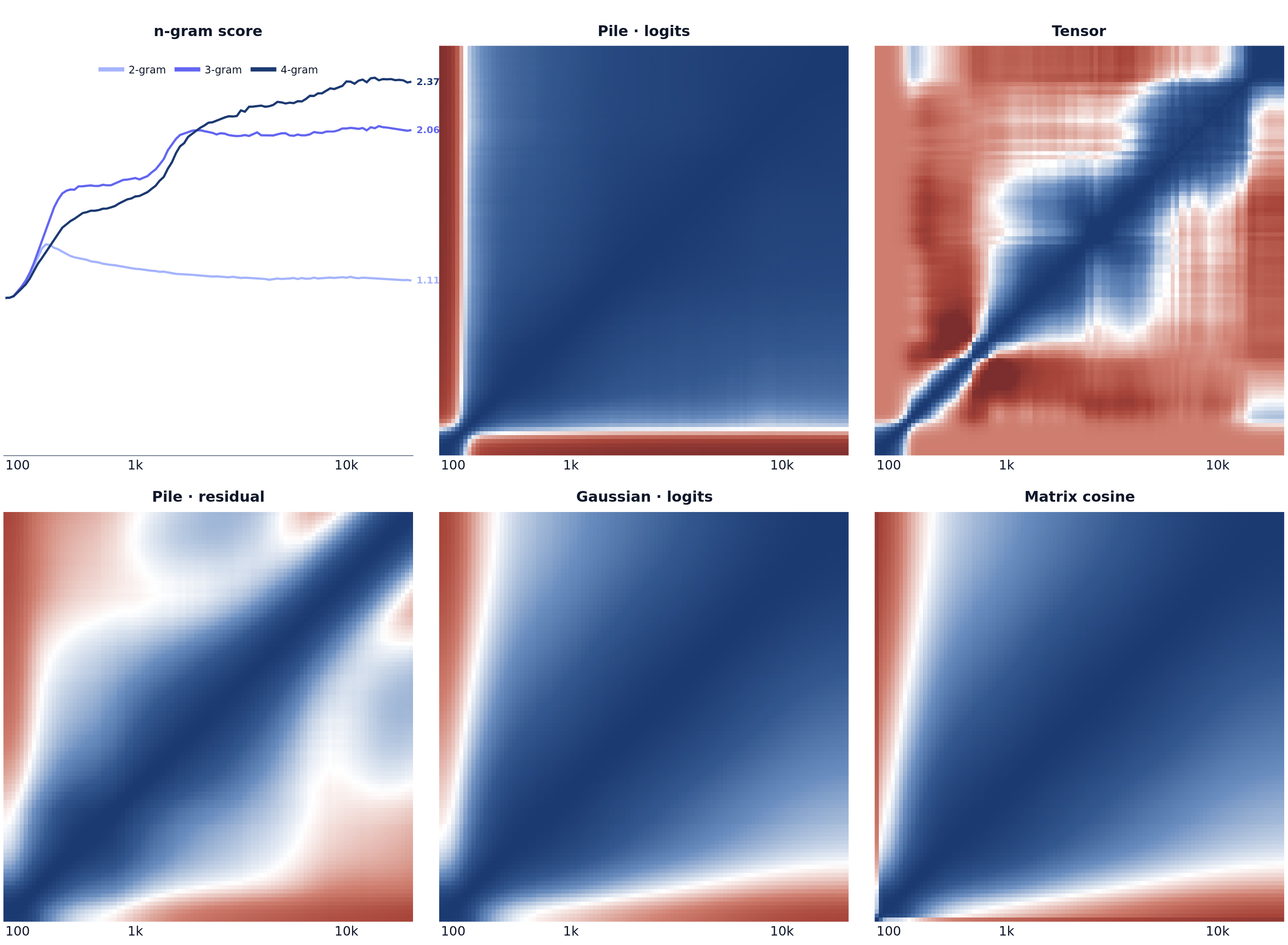}
    \caption{Five similarity measures across 101 log-spaced checkpoints of a two-layer bilinear attention transformer on The Pile. We track the n-gram score as a heuristic measure of change, which shows continued changes throughout most of training. Both the Pile $\cdot$ residual and the Pile $\cdot$ logits are behavioural similarities on pre-unembedding activations and on logits over Pile tokens; Gaussian $\cdot$ logits is the same on logits driven by matched-$\sigma$ Gaussian inputs. Tensor is the closed-form weight-based similarity; Matrix cosine is Frobenius cosine of flattened weights, as discussed in \autoref{fig:Motivation}.}
    \label{fig:language_similarity}
\end{figure}

The previous results used 1-layer bilinear models. Tensor similarity scales unchanged to deeper transformers and continues to find structure where existing metrics blur. We train a two-layer bilinear attention transformer on The Pile~\citep{ThePile} and save 101 log-spaced checkpoints, with further setup details reported in~\autoref{app:language}. \autoref{fig:language_similarity} reports five similarity measures across these checkpoints alongside the n-gram behaviour of the model, which prior work uses to track the increasingly complex structures models acquire over training, often discontinuously \citep{wang2024differentiationspecializationattentionheads, nguyen2024understandingtransformersngramstatistics, belrose2024neuralnetworkslearnstatistics}. Existing metrics separate only the earliest checkpoints from the rest, blurring everything beyond; tensor similarity exhibits clear block structure with sharp transitions between regimes.

More broadly, behavioural metrics see only the slice of input space they are evaluated on, and a mechanism that has reorganised globally need not register on that slice if its predictions there happen to coincide \citep{damour2020underspecificationpresentschallengescredibility}. What generalises is what holds beyond the training distribution; measuring similarity on the training distribution conflates it with memorisation. Tensor similarity integrates over the full Gaussian input space and so registers reorganisations that would be invisible to any finite probe, which is precisely the regime where mechanistic claims need to hold.

\section{Conclusion}\label{sec:conclusion}
We introduce tensor similarity, a principled metric for evaluating functional equivalence from the weights of tensor-based models such as bilinear transformers. We then apply our metric to problems in mechanistic and developmental interpretability.

\paragraph{Summary.}\label{summary}
Measuring functional similarity from weights requires accounting for the symmetries that leave a model's outputs invariant. Tensor similarity projects these out; combined with the polarisation isomorphism, the result is a principled measure of functional equivalence. Exploiting the link to behavioural similarity under Gaussian inputs, we further derive a Gaussian metric that yields a stricter equivalence relation. We validate the framework on four tasks: catastrophic forgetting in vision, backdoor insertion via dataset poisoning, grokking in modular arithmetic, and the formation of $n$-gram statistics during language modelling on The Pile. Across all four tasks, tensor similarity tracks functional changes more cleanly than existing metrics, and slicing localises that change to specific outputs or components.

\paragraph{Limitations.}\label{limitations}
The primary limitation is the reliance on multilinear models. This excludes most state-of-the-art architectures where non-polynomial non-linearities between components block the algebraic computation of functional properties from weights, particularly for cross-layer mechanisms. This limitation bites hardest if multilinear models cannot match standard architectures in performance, leaving no incentive to adopt them at scale. Even in that case, we believe multilinear models earn their place as test beds for mechanistic interpretability — they form a non-trivial class of architectures that admits effective algebraic analysis — and regardless, there is mounting evidence that multilinear models can be in the same league as state-of-the-art architectures \citep{pearce2025bilinear, shazeer2020gluvariantsimprovetransformer}. A secondary limitation is the limited scope of the analysis conducted here. While theory and the diff construction gesture towards attribution, our experiments demonstrate detection of functional changes, such as backdoors, more thoroughly than they demonstrate attribution of the specific model components that cause them. 


\paragraph{Outlook.}\label{outlook}
A natural next step toward the goal of attribution, a key piece of the ambitious mechanistic interpretability agenda, is local application: computing tensor similarity between paths through a deep multilinear transformer and extending the framework of \citep{elhage2021mathematical}. This could include bilinear MLPs and reveal mechanisms beyond the two-layer attention-only case that originally produced induction heads. The flipside of the restriction to multilinear models is that this work adds to a growing literature showing they are much more transparent than ordinary deep networks. Their symmetries are well understood, polarisation identifies a function from its weights alone, and norms and correlation measures can be traced globally through the weights \citep{levineQuantumEntanglementDeep2019}. Combined with evidence that multilinear models match non-linear ones on standard benchmarks \citep{pearce2025bilinear, cohenConvolutionalRectifierNetworks2016, riggs2026tensorTransformerVariants}, the case builds for multilinear architectures as a foundation for trustworthy machine learning. 



\section*{Acknowledgements}

This work was supported by the MARS (Mentorship for Alignment Research Students) program run by the Cambridge AI Safety Hub (\url{caish.org/mars}). We are grateful to the people of CAISH for making our collaboration possible, and to our research manager, Mikhail Mironov, for ensuring we didn't veer too far off track. Additionally, we thank Ward Gauderis, Richard Mohn, and Viktor Rehnberg for technical discussions and encouragement.


\bibliographystyle{plainnat}
\bibliography{references}

\clearpage
\appendix

\section{Theory appendix}\label{app:theory}

\subsection{Multilinear models and their symmetries.}
\label{app:theory_A}

\paragraph{Multilinear models.}
Multilinear models take the following form: $\mathcal A(\mathbf x)=\mathbf A\cdot\otimes^{n}\tilde{\mathbf x}$ where $\mathbf x\in\mathbb R^d$, $\tilde{\mathbf x}=(1,\mathbf x)$, $\mathbf A\in\mathbb R^K\otimes^n\mathbb R^{d+1}$, and $\cdot$ denotes the tensor contraction along $\otimes^n\mathbb R^{d+1}$. In component form:
$\mathcal A(\mathbf x)_k=\sum_{\mathbf i\in[d]^n_0}A_{ki_1...i_n}x_{i_1}\cdots x_{i_n}$, which is a polynomial of order $n$ in the components of $\mathbf x$. $\mathcal A$ thus models $K\in\mathbb N$ polynomials.

\paragraph{Deep learning with multilinear models.}
The order of the polynomial governs the expressiveness of multilinear models.
However, the size of the tensor $\mathbf A$ grows exponentially with order as $\mathcal{O}(K(d+1)^n)$: this is the curse of dimensionality.

To tackle this with multilinear models, we layer them. Consider a deep multilinear model with $l\in\mathbb N$ layers:
\[
\mathcal A(\mathbf x)=\mathcal A^{(l)}(\mathcal A^{(l-1)}\cdots(\mathcal A^{(1)}(\mathbf x))\cdots)=\mathbf A^{(l)}\cdot \otimes^{n_l}(\mathbf A^{(l-1)}\cdots\otimes^{n_2}(\mathbf A^{(1)}\cdot\otimes^{n_1}\tilde{\mathbf x})),
\]
where $n_i\in\mathbb N$ denotes the input order of the local tensor $\mathbf A^{(i)}\in\mathbb R^{H_i}\otimes^{n_i}\mathbb R^{H_{i-1}}$ for $i\in[l]$ and $\{H_i\}$ the hidden dimensions with $H_{0}=d+1$ and $H_l=K$. While the global tensor has order $n=1+\prod_{i=1}^ln_i$, only local tensors of order $n_i$ need to be stored and this therefore constitutes a tensor decomposition of the global tensor $\mathbf A$. 

Tensor networks are decompositions of tensors according to an underlying graph. The graph that underlies the decomposition of a deep multilinear model has tree shape, with the tensors within each layer being copies of each other. Thus, while the order of the modelled tensor is exponential in the number of layers, the number of coefficients that need to be stored in the local tensors grows only linearly with the number of layers.

\paragraph{Action of the permutation group.}
Consider a multilinear model given by the tensor $\mathbf A\in\mathbb R^K\otimes^n\mathbb R^{d+1}$ and the symmetric group on $n$ elements, $\mathcal S_n$. Its action on $\mathbb R^K\otimes^n\mathbb R^{d+1}$ permutes the copies $\otimes^n\mathbb R^{d+1}$ as $A_{ki_1\dots i_n}\to A_{k i_{\sigma(1)}\dots i_{\sigma(n)}}$. Tensors that are invariant under this action are said to be symmetric. The space of these symmetric tensors, $\mathbb R^K\otimes \mathrm{sym}^n\mathbb R^{d+1}$, is a proper vector subspace of the vector space of generic tensors. The projection down to the symmetric subspace, called symmetrisation and denoted by $\mathbf P_n:\otimes^n\mathbb R^{d+1}\to\mathrm{sym}^n\mathbb R^{d+1}$, is a linear orthogonal projection defined as the mean of all the actions of the symmetric group:
\begin{equation}
\mathbf A^{\mathrm{sym}}:=(\mathds 1\otimes\mathbf P_n)\cdot\mathbf A,\hspace{1em}A^{\mathrm{sym}}_{ki_1 \cdots i_n}
  \;=\;
  \frac{1}{n!} \sum_{\sigma \in \mathcal{S}_n} A_{ki_{\sigma(1)} \cdots i_{\sigma(n)}}.
\end{equation}
The orthogonal projector $\mathbf{P}_n$ satisfies $\mathbf{P}_n^2 = \mathbf{P}_n$ and
$\langle \mathbf{P}_n \mathbf{A} \mid \mathbf{B} \rangle =
\langle \mathbf{A} \mid \mathbf{P}_n \mathbf{B} \rangle$.

\paragraph{Polarisation.}
This symmetry leaves the function $\mathcal A$ invariant, and so permuting the input copies leaves the modelled polynomial invariant. The precise statement of this is the polarisation isomorphism, usually stated for homogeneous polynomials. It states that there is a vector space isomorphism between the symmetric tensors over $\mathbb R^d$ of order $n$, $\mathrm{sym}^n\mathbb R^d$, and the space of homogeneous polynomials of order $n$ in $\mathbf x\in\mathbb R^d$. In our case, because we consider general (not merely homogeneous) polynomials, we lift the input to $\tilde{\mathbf x}=(1,\mathbf x)$ and work with degree-$n$ homogeneous polynomials over $\mathbb R^{d+1}$; by polarisation these are in bijection with $\mathrm{sym}^n\mathbb R^{d+1}$. This applies for each slice $\mathbf A_k\in\mathrm{sym}^n\mathbb R^{d+1}$ of the multilinear model considered. The output space $\mathbb R^K$ has a privileged basis such that slicing along it is principled. For a single-layer multilinear model, therefore, symmetrising the input indices of the tensor $\mathbf A$ puts it into a one-to-one correspondence with the vector of polynomials modelled, accounting for all weight-space symmetries.

\paragraph{Example: bilinear layer.}
A \emph{bilinear layer} is a single-layer multilinear model of order $n=2$, so its weight tensor $\mathbf{A}\in\mathbb{R}^K\otimes^2\mathbb{R}^{d+1}$ encodes $K$ quadratic polynomials. Given matrices $\mathbf{D}\in\mathbb{R}^{K\times r}$ and $\mathbf{L},\mathbf{R}\in\mathbb{R}^{r\times(d+1)}$ with hidden dimension $r\in\mathbb{N}$, and the order-3 delta tensor $\boldsymbol{\delta}\in\mathbb{R}^r\otimes\mathbb{R}^r\otimes\mathbb{R}^r$ with components $\delta_{h_1h_2h_3}=1$ iff $h_1=h_2=h_3$ and zero otherwise, the weight tensor is
\begin{equation}\label{eq:bilinear_cp}
  A_{k i_1 i_2}
  = \sum_{h_1,h_2,h_3=1}^{r} D_{kh_1}\,\delta_{h_1h_2h_3}\,L_{h_2 i_1}\,R_{h_3 i_2}
  = \sum_{h=1}^{r} D_{kh}\,L_{h i_1}\,R_{h i_2}.
\end{equation}
This is a rank-$r$ CP decomposition, equivalently written $\mathbf{A}=\sum_{h=1}^r\mathbf{d}_h\otimes\mathbf{l}_h\otimes\mathbf{r}_h$, where $\mathbf{d}_h,\mathbf{l}_h,\mathbf{r}_h$ are the $h$-th columns of $\mathbf{D}$ and rows of $\mathbf{L},\mathbf{R}$ respectively. The forward pass reduces to
\begin{equation}
  \mathcal{A}(\tilde{\mathbf{x}}) = \mathbf{D}\bigl[(\mathbf{L}\tilde{\mathbf{x}})\odot(\mathbf{R}\tilde{\mathbf{x}})\bigr],
\end{equation}
where $\odot$ denotes the Hadamard (elementwise) product, computable with matrix operations alone.

The output depends only on the symmetric part of $\mathbf{A}$.  Decompose
$\mathbf{A} = \mathbf{A}^{\mathrm{sym}} + \mathbf{A}^{\mathrm{asym}}$, where
$A^{\perp}_{ki_1i_2} = \tfrac{1}{2}(A_{ki_1i_2} - A_{ki_2i_1})$ is the antisymmetric
complement.  Since $\tilde{\mathbf{x}}\otimes\tilde{\mathbf{x}}$ is symmetric under
$i_1\leftrightarrow i_2$, the contraction with $\mathbf{A}^{\mathrm{asym}}$ vanishes for all
$\tilde{\mathbf{x}}$:
\[
  \sum_{i_1,i_2} A^{\mathrm{asym}}_{ki_1i_2}\,\tilde{x}_{i_1}\tilde{x}_{i_2}
  \;=\;
  \tfrac{1}{2}\sum_{i_1,i_2}(A_{ki_1i_2}-A_{ki_2i_1})\tilde{x}_{i_1}\tilde{x}_{i_2}
  \;=\; 0.
\]
Therefore $\mathcal{A}(\tilde{\mathbf{x}}) = \mathbf{A}^{\mathrm{sym}}\cdot_{\mathrm{in}}(\tilde{\mathbf{x}}\otimes\tilde{\mathbf{x}})$,
and in terms of the CP factors,
\begin{equation}\label{eq:bilinear_sym}
  A^{\mathrm{sym}}_{ki_1i_2}
  \;=\;
  \tfrac{1}{2}\sum_{h=1}^{r} D_{kh}\bigl(L_{hi_1}R_{hi_2} + R_{hi_1}L_{hi_2}\bigr).
\end{equation}
Two bilinear layers with different $(\mathbf{L},\mathbf{R})$ but equal $\mathbf{A}^{\mathrm{sym}}$
implement the same polynomial; weight-based comparison is therefore only principled after
symmetrisation, as established by the polarisation isomorphism above.

\paragraph{The symmetry of deep multilinear models}
Layering multilinear models introduces an additional subtlety. It is too expensive to instantiate the full tensor $\mathbf A$ that decomposes according to a tree into the local tensors $\mathbf A^{(i)}$ for $i\in[l]$. However, full permutations of all input indices of the global tensor would require instantiating the full tensor. Moreover, these full permutations would destroy the tree decomposition structure. Because leaving the decomposition space is computationally intractable, we do not consider the action of the full symmetric group. Restricting to these tree decompositions of the global tensor, we identify the symmetry group that is compatible with the tree decomposition of $\mathbf A$ to be the wreath product of local permutations: $G_\mathrm{tree}:=\mathcal S_{n_l}\wr\mathcal S_{n_{l-1}}\wr\cdots\wr\mathcal S_{n_1}\subsetneq\mathcal S_n$. Symmetrising with respect to this group symmetrises each layer in turn, resulting in the locally symmetrised $\mathbf A^{(l),\mathrm{sym}}\cdot \otimes^{n_l}(\mathbf A^{(l-1),\mathrm{sym}}\cdots\otimes^{n_2}(\mathbf A^{(1),\mathrm{sym}}))$.
Projecting $\mathbf{A}$ to the symmetric subspace of $G_\mathrm{tree}$ by symmetrising each layer locally removes all degrees of freedom related to input permutation symmetry, while remaining within the class of multilinear models sharing the same tree decomposition. This enables principled weight-based comparison between models of the same decomposition class.

Another symmetry of layered multilinear models concerns basis transformations of the hidden dimensions along which the different layers are connected. This is discussed next but concerns only functions of local tensors (not of the global tensor).

\paragraph{Gauge symmetry.} For tensor decompositions, just like matrix decompositions, the exact values of the local tensors are not unique. Inserting $\mathbf{U}\mathbf{U}^{-1}$ in any contracted dimension leaves the global tensor invariant. Functions of the global tensor, including the modelled polynomial for multilinear models or the inner product between two decomposed tensors, are also invariant under gauge transformations. Gauge fixing is the procedure to remove these degrees of freedom. For our models, privileged output and input bases, as well as basis-dependent computations such as the elementwise product of tensors, fix a gauge. Combining this with a residual stream that propagates the privileged input basis through the layers of a deep multilinear model fixes a gauge. This is only relevant if the tensors of individual layers are analysed instead of computing functions from the global tensor such as tensor similarity.

\paragraph{Implications for tensor similarity.}
The central choice in defining tensor similarity is the metric tensor $\mathbf{M} \in \otimes^2(\otimes^n \mathbb{R}^{d+1})$. Three requirements constrain this choice: it should project out weight-space symmetries, relate deviation from unity with functional difference like behavioural similarity, and preserve the class of deep multilinear models, i.e.\ those whose global tensors admit a special tree decomposition.

When the metric projects out weight-space symmetries (for instance, when combined with layerwise symmetrisation), tensor similarity reduces to the inner product between two tensors that are in one-to-one correspondence with polynomials. The Cauchy--Schwarz inequality relates this inner product to the product of the tensor norms, precisely the quantities appearing in the denominator of tensor similarity. It further guarantees that tensor similarity equals one if and only if the tensors, and hence the polynomials, are related by a positive scalar.
\subsection{Gram recursion: derivation, symmetrisation compatibility, and algorithms}
\label{app:theory_C}


This appendix provides the additional details used in Section~\ref{sec:global_weight_based}. Here we:
\begin{itemize}
  \item derive the Gram recursion from the tree structure of layered networks;
  \item prove the commutativity result: $\mathbf{P}_{n_i}$ commutes with the Gram step, so layer-wise symmetrisation is compatible with the recursion;
  \item give the explicit four-matrix-product algorithm for bilinear layers ($n_i = 2$) and state its complexity.
\end{itemize}

Define the Gram matrix at layer $i$ as the partial contraction
\begin{equation}
  \label{eq:gram_recursion_app}
  \mathbf{G}^{(i)}
  \;=\;
  \mathbf{A}^{(i)} \cdot_{\mathrm{in}}
  \bigl(\mathbf{G}^{(i-1)}\bigr)^{\otimes q_i} \cdot_{\mathrm{in}}
  \mathbf{B}^{(i)},
  \qquad \mathbf{G}^{(0)} = \mathbf{I}.
\end{equation}
The inner product is recovered as $\langle \mathbf{A} \mid \mathbf{B} \rangle =
\mathrm{tr}\,\mathbf{G}^{(L)}$.

\textbf{Derivation of the recursion.}
The global weight tensor $\mathbf{A}$ of a layered network is a tree tensor network whose
leaves are the local tensors $\mathbf{A}^{(1)},\ldots,\mathbf{A}^{(L)}$.  The inner
product $\langle\mathbf{A}\mid\mathbf{B}\rangle$ contracts all indices of $\mathbf{A}$
against the corresponding indices of $\mathbf{B}$.  Because the underlying tree is
acyclic, contraction can proceed from the leaves upward: contracting the input legs
of layer $i$ uses only the local tensors $\mathbf{A}^{(i)},\mathbf{B}^{(i)}$ and
the result of the contraction at layer $i-1$, which is captured by $\mathbf{G}^{(i-1)}$.
Initialising with $\mathbf{G}^{(0)}=\mathds{1}\in\otimes^2\mathbb{R}^{d+1}$ (the identity
on the input space) and applying this observation recursively yields
\eqref{eq:gram_recursion_app}.  The inner product is the trace of the final Gram matrix
because the remaining open legs at layer $L$ are the output legs, and tracing them gives
$\langle\mathbf{A}\mid\mathbf{B}\rangle=\sum_k G^{(L)}_{kk}=\mathrm{tr}\,\mathbf{G}^{(L)}$.
The next paragraph shows a commutativity result that reduces the computational cost of layerwise symmetrisation during the Gram recursion.
The final paragraph of this subsection gives the algorithm for the computation of the Gram recursion explicitly for bilinear layer. 

\textbf{Commutativity.}
Let $\mathbf{G}\in\otimes^2\mathbb R^H$ and $\mathbf{P}_{n}$ the
symmetrisation projector on $\bigotimes^{n}\mathbb R^H$.  Then
$[\mathbf{P} _{n},\,\otimes^{n}\mathbf{G}]=0$.

\textit{Proof.}
It suffices to show that each permutation operator $\mathbf{U}_\sigma$,
$\sigma\in\mathcal{S}_n$, commutes with $\otimes^{n}\mathbf{G}$, since
$\mathbf{P}_n$ is a linear combination of these.  On product vectors,
\[
  \mathbf{U}_\sigma\,\otimes^{n}\mathbf{G}(\mathbf x_1\otimes\cdots\otimes \mathbf x_n)
  = \mathbf G\mathbf x_{\sigma^{-1}(1)}\otimes\cdots\otimes\mathbf  G\mathbf x_{\sigma^{-1}(n)}
  = \otimes^{n}\mathbf{G}\,\mathbf{U}_\sigma(\mathbf x_1\otimes\cdots\otimes \mathbf x_n).
\]
Product vectors span $\bigotimes^{n}\mathbb R^H$, so the claim follows.
\hfill$\square$

By idempotency of $\mathbf{P}_{n_i}$ and the above commutativity result,
$\mathbf{P}_{n_i}\cdot\otimes^{n_i}\mathbf{G}^{(i-1)}\cdot
\mathbf{P}_{n_i}=\otimes^{n_i}\mathbf{G}^{(i-1)}\cdot\mathbf{P}_{n_i}$.
It follows that the symmetrised Gram recursion requires symmetrising only one of the two
layers per step:
\begin{equation}\label{eq:sym_gram_recursion_app}
  \mathbf{G}^{(i),\mathrm{sym}}
  \;=\;
  \mathbf{A}^{(i)}\cdot_{\mathrm{in}}
  \bigl(\otimes^{n_i}\mathbf{G}^{(i-1),\mathrm{sym}}\bigr)
  \cdot_{\mathrm{in}}\mathbf{P}_{n_i}\mathbf{B}^{(i)}.
\end{equation}

\textbf{Bilinear algorithm.}
For bilinear layers ($n_i=2$) parameterised by
$\mathbf{L}_\mathcal{A}^{(i)},\mathbf{R}_\mathcal{A}^{(i)}\in\mathbb{R}^{r_i\times H_{i-1}}$,
$\mathbf{D}_\mathcal{A}^{(i)}\in\mathbb{R}^{H_i\times r_i}$ (and likewise for $\mathcal{B}$),
the Gram step \eqref{eq:gram_recursion_app} reduces to matrix operations alone.
Initialise $\mathbf{G}^{(0)}=\mathds{1}_{d+1}$ and, for each $i\in[L]$, compute four
matrix products,
\begin{align}
  (\mathbf{LL})^{(i)} &:= \mathbf{L}_\mathcal{A}^{(i)}\,\mathbf{G}^{(i-1)}\,
    \bigl(\mathbf{L}_\mathcal{B}^{(i)}\bigr)^T, &
  (\mathbf{RR})^{(i)} &:= \mathbf{R}_\mathcal{A}^{(i)}\,\mathbf{G}^{(i-1)}\,
    \bigl(\mathbf{R}_\mathcal{B}^{(i)}\bigr)^T, \notag\\
  (\mathbf{LR})^{(i)} &:= \mathbf{L}_\mathcal{A}^{(i)}\,\mathbf{G}^{(i-1)}\,
    \bigl(\mathbf{R}_\mathcal{B}^{(i)}\bigr)^T, &
  (\mathbf{RL})^{(i)} &:= \mathbf{R}_\mathcal{A}^{(i)}\,\mathbf{G}^{(i-1)}\,
    \bigl(\mathbf{L}_\mathcal{B}^{(i)}\bigr)^T,
  \label{eq:bilinear_gram_four}
\end{align}
two elementwise products,
\begin{align}
  \mathbf{E}_{\|}^{(i)} &:= (\mathbf{LL})^{(i)}\odot(\mathbf{RR})^{(i)}, &
  \mathbf{E}_{\times}^{(i)} &:= (\mathbf{LR})^{(i)}\odot(\mathbf{RL})^{(i)},
  \label{eq:bilinear_gram_hadamard}
\end{align}
and one final matrix multiplication,
\begin{equation}\label{eq:bilinear_gram_final}
  \mathbf{G}^{(i)}
  = \mathbf{D}_\mathcal{A}^{(i)}
    \bigl(\mathbf{E}_{\|}^{(i)}+\mathbf{E}_{\times}^{(i)}\bigr)
    \bigl(\mathbf{D}_\mathcal{B}^{(i)}\bigr)^T.
\end{equation}
The two terms $\mathbf{E}_\|$ and $\mathbf{E}_\times$ correspond to the two terms in the
symmetrisation of the bilinear layer: $\mathbf{E}_\|$ pairs same-role legs ($L$--$L$ and
$R$--$R$) while $\mathbf{E}_\times$ pairs swapped-role legs ($L$--$R$ and $R$--$L$).
Our experiments are implemented in Python with the tensor contraction algorithms like the above implemented using \texttt{quimb}~\cite{gray2018quimb}.
\subsection{Derivation of the Gaussian metric}
\label{app:theory_D}


This appendix derives the Gaussian metric tensor presented in Section~\ref{sec:symmetries_and_metric}. Here we:
\begin{itemize}
  \item show that the expected output similarity equals $\langle \mathbf{A} \mid \mathbf{M}^{(p)} \mid \mathbf{B} \rangle$ where $\mathbf{M}^{(p)}$ has $\mathcal{S}_{2n}$ symmetry;
  \item compute $\boldsymbol{\Lambda}^{(p)}$ for a Gaussian input distribution via the Isserlis--Wick theorem;
  \item conclude that this metric projects onto a strictly smaller subspace than the plain $\mathcal{S}_n$ symmetrisation.
\end{itemize}

Taking the expectation over inputs of the activation inner product gives
\begin{equation}
  \label{eq:s2n_metric_start}
  \mathbb{E}_{\mathbf{x}}\bigl[\langle \mathcal{A}(\mathbf{x}) \mid
  \mathcal{B}(\mathbf{x})\rangle\bigr]
  \;=\;
  \langle \mathbf{A} \mid \mathbf{M}^{(p)} \mid \mathbf{B} \rangle,
\end{equation}
where $\mathbf{M}^{(p)} = \mathds{1} \otimes \boldsymbol{\Lambda}^{(p)}$ and
\begin{equation}
  \boldsymbol{\Lambda}^{(p)}_{i_1 \cdots i_n j_1 \cdots j_n}
  \;=\;
  \mathbb{E}_{\mathbf{x}}\bigl[x_{i_1} \cdots x_{i_n} x_{j_1} \cdots x_{j_n}\bigr].
\end{equation}
The tensor $\boldsymbol{\Lambda}^{(p)}$ has order $2n$, pairing all $2n$ input legs of
$\mathbf{A}$ and $\mathbf{B}$, including self-contractions of two legs belonging to the
same tensor. This makes $\boldsymbol{\Lambda}^{(p)}$ manifestly invariant under the action
of $\mathcal{S}_{2n}$.

\textbf{Gaussian case via Isserlis' theorem.}
For $\mathbf{x} \sim \mathcal{N}(0, \mathbf{I}_d)$, Isserlis'
theorem~\citep{munthe-kaasShortProofIsserlis2025} expresses the $2n$-th moment as a sum
over all perfect matchings of the $2n$ indices $i_1,\ldots,i_n,j_1,\ldots,j_n$.
For unit covariance, each matched pair $(a,b)$ contributes $\delta_{i_a i_b}$.

Classify pairings by $m\in\mathbb N$, the number of internal $\mathbf{A}$-pairs (pairs drawn entirely from
$\{i_1,\ldots,i_n\}$). A pairing with $m$ internal $\mathbf{A}$-pairs must also have $m$ internal
$\mathbf{B}$-pairs (because the total number of $\mathbf{A}$-indices must equal the number of $\mathbf{B}$-indices), leaving $n-2m$ cross-pairs.

Define the partial trace operator $\tau$ such that $\tau^m \mathbf{T}$ contracts $m$ pairs of indices as
\begin{equation}
  (\tau\, \mathbf{T})_{i_3\cdots i_n}
  \;=\;
  \sum_{a=1}^d T_{a\,a\,i_3\cdots i_n}.
\end{equation}
All pairings of type $m$
contribute the same traced inner product $\langle \tau^m\mathbf{A}|\, \tau^m\mathbf{B}\rangle$.

Counting the pairings of type $m$: on the $\mathbf{A}$-side, choose and pair $2m$ of the $n$
indices in $\binom{n}{2m}(2m-1)!!$ ways; the same count applies to $\mathbf{B}$.
The remaining $n-2m$ legs on each side are then cross-matched in $(n-2m)!$ ways.
Setting
\begin{equation}
  c_{n,m} \;=\; \binom{n}{2m}^2\,(2m-1)!!^{\,2}\,(n-2m)!,
\end{equation}
and summing over $m$ gives the main result:
\begin{equation}
  \label{eq:gaussian_sim_homogeneous}
  \mathbb{E}_{\mathbf{x}\sim\mathcal{N}(0,\mathbf{I}_d)}
  \bigl[\langle\mathcal{A}(\mathbf{x})\mid\mathcal{B}(\mathbf{x})\rangle\bigr]
  \;=\;
  \sum_{m=0}^{\lfloor n/2\rfloor}
  c_{n,m}\,
  \langle \tau^m\mathbf{A},\, \tau^m\mathbf{B}\rangle.
\end{equation}


Equation~\eqref{eq:gaussian_sim_homogeneous} identifies
$\boldsymbol{\Lambda}^{(p)}$ as a weighted sum of partial-trace operators, confirming that
$\mathbf{M}^{(p)}$ is positive definite and carries full $\mathcal{S}_{2n}$ symmetry.
Because the $\mathcal{S}_{2n}$-invariant subspace of $\bigotimes^n\RR^{d+1}$ is strictly
smaller than the $\mathcal{S}_n$-invariant one, this Gaussian metric is a strictly finer
equivalence relation than plain $\mathcal{S}_n$-symmetrisation, while still satisfying the
discriminative-power guarantee \eqref{eq:sym_sim_discriminative}.

\section{Additional catastrophic forgetting details}\label{app:cata_forget}

\textbf{Dataset:} The SVHN dataset comprises over $600,000$ labelled digits cropped from Google Street View images~\citep{SVHN}. It occupies a useful middle ground in difficulty: unlike MNIST~\citep{deng2012mnist}, it is not solvable using linear models, yet it does not require the large architectures for ImageNet-scale tasks. We use a subset of approximately $73,000$ training and $26,000$ validation images, converted to greyscale. 

\textbf{Model Architecture:}~The model consists of a linear embedding layer,  a single bilinear layer, and a linear unembedding layer. For ease of analysis, biases and normalisation layers are excluded from the model.

\textbf{Model Dimensions and Training:}~Table~\ref{tab:svhn-params} summarises the model dimensions and training hyperparameters.

\begin{table}[htbp]
    \centering
    \begin{tabular}{lr}
        \toprule
        \multicolumn{2}{c}{\textbf{Architecture}} \\
        \midrule
        \textbf{d\_model}  & 128 \\
        \textbf{d\_hidden} & 256 \\
        \textbf{n\_layer}  & 1 \\
        \midrule
        \multicolumn{2}{c}{\textbf{Training Parameters}} \\
        \midrule
        \textbf{dropout}       & 0.0 \\
        \textbf{weight decay}  & 0.5  \\
        \textbf{batch size}    & 248 \\
        \textbf{learning rate} & $10^{-3}$ \\
        \textbf{optimizer}     & AdamW \\
        \textbf{schedule}      & Cosine Annealing \\
        \textbf{epochs}        & 20 per stage \\
        \bottomrule
    \end{tabular}
    \caption{Model architecture and training setup for the catastrophic forgetting setup.}
    \label{tab:svhn-params}
\end{table}

\begin{figure}[tbh!]
    \centering
    \includegraphics[width=\linewidth]{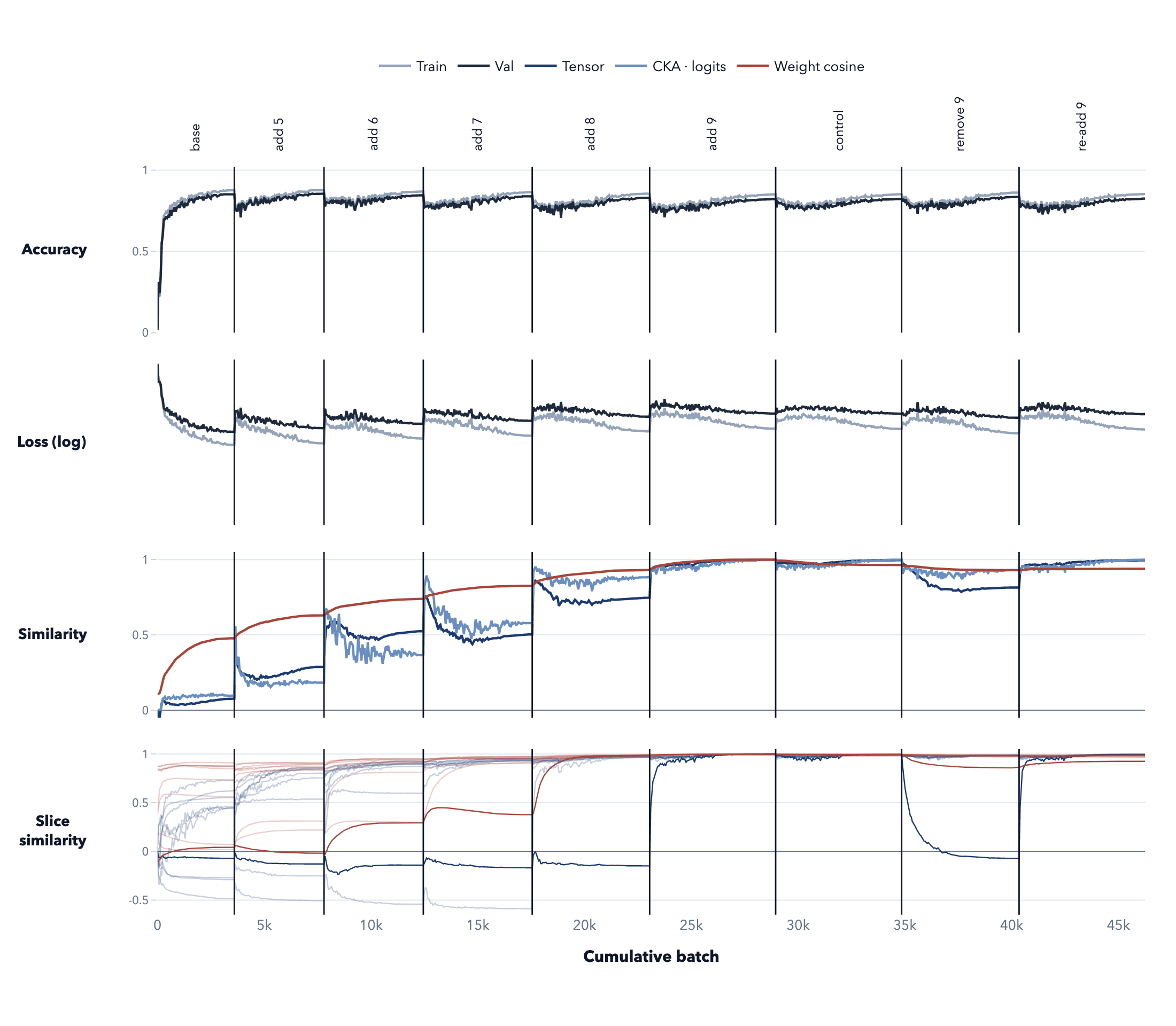}
    \caption{Model evolution across the progressive training setup on SVHN described in \autoref{sec:cataforget}. \textbf{Accuracy and loss} follow a consistent pattern across all stages, with validation accuracy reaching $83$--$85$\% at each stage. \textbf{Similarity} with respect to the model at the end of  the \textit{add 9} stage is shown for tensor similarity, CKA on logits, and weight cosine similarity. All three measures increase as digits are incrementally added, but tensor similarity exhibits the most pronounced drop at the \textit{remove 9} stage and the clearest recovery at \textit{re-add 9}, while CKA and weight cosine show considerably weaker signals. \textbf{Slice similarity} highlights digit `9' in a darker line and all other classes shown faint. The drop at \textit{remove 9} is localised almost entirely to digit `9', confirming that catastrophic forgetting affects this class specifically. The effect is very clearly captured by tensor slice similarity, and is significantly weaker in weight cosine slice similarity.}
    \label{fig:app_svhn_progress}
\end{figure}

\begin{figure}[tbh!]
    \centering
    \includegraphics[width=\linewidth]{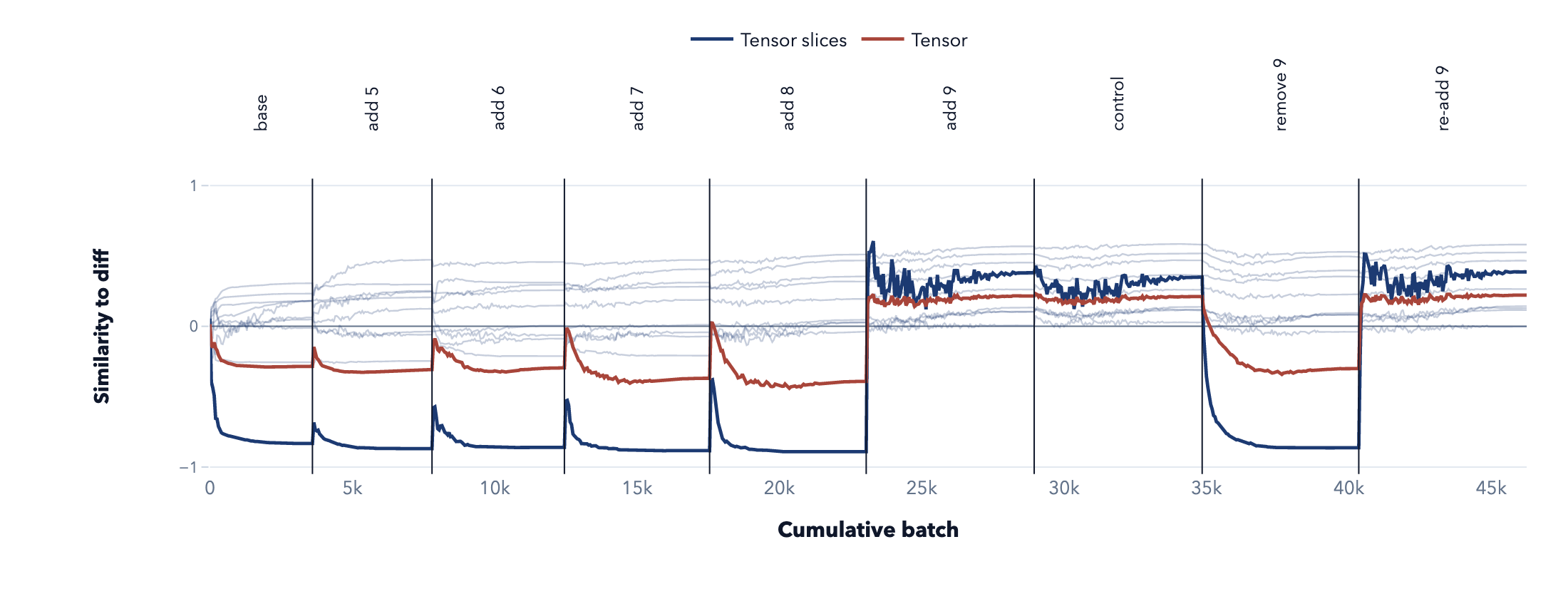}
    \caption{Tensor diff similarities across the progressive SVHN training setup (\autoref{sec:cataforget}). \textbf{Similarity to diff} is computed with respect to a diff tensor, computed as the difference between two checkpoints of an identically-architectured bilinear model trained on a different subset of SVHN data with a different random seed. The first checkpoint is taken after training on  $\{0,\cdots,8\}$ and the second after subsequent fine-tuning on the full set  $\{0,\cdots,9\}$, thereby isolating weight updates associated with learning digit `9'. Tensor similarity is shown in red and slice similarities are shown in blue, with digit `9' highlighted in a darker line. The resulting pattern closely mirrors the self-similarity result of \autoref{fig:app_svhn_progress} with catastrophic forgetting at \textit{remove 9} and subsequent recovery at \textit{re-add 9} clearly  visible, demonstrating that an externally constructed diff tensor can isolate the same functional component as self-comparison.}
    \label{fig:app_svhn_diff}
\end{figure} 
\section{Additional modular addition details}
\label{app:mod_arith_extra}

\textbf{Dataset:}~The task is modular addition: given inputs $(a, b)$, predict $(a + b) \bmod p$ where $p = 113$. Inputs are represented as two one-hot vectors of dimension $2p = 226$, concatenating the encodings of $a$ and $b$. All $p^2 = 12{,}769$ input pairs are used, with a $60:40$ split between train and validation sets.

\textbf{Model Architecture:}~The model consists of a single bilinear layer followed by a linear unembedding layer, with biases and normalisation layers.

\textbf{Model Dimensions and Training:}~Table~\ref{tab:grokking-params} summarises the model dimensions and training hyperparameters.

\begin{table}[htbp!]
    \centering
    \begin{tabular}{lr}
        \toprule
        \multicolumn{2}{c}{\textbf{Architecture}} \\
        \midrule
        \textbf{d\_input}  & 226 \\
        \textbf{d\_hidden} & 64 \\
        \textbf{n\_layer}  & 1 \\
        \midrule
        \multicolumn{2}{c}{\textbf{Training Parameters}} \\
        \midrule
        \textbf{dropout}       & 0.0 \\
        \textbf{weight decay}  & 0.06 \\
        \textbf{batch size}    & 512 \\
        \textbf{learning rate} & $10^{-3}$ \\
        \textbf{optimizer}     & AdamW \\
        \textbf{schedule}      & constant \\
        \textbf{steps}         & 100{,}000 \\
        \bottomrule
    \end{tabular}
    \caption{Model architecture and training setup for the modular addition experiment.}
    \label{tab:grokking-params}
\end{table}

\section{Language modelling with an attention only transformer}
\label{app:language}

\subsection{Training details}
\label{app:language_training}

\subsubsection{Data and tokenisation}

We train on the DSIR-filtered-Pile-50M subset from~\cite{xie2023data}. Validation is created using a deterministic 1\% held-out document split from the training set. We train a custom BPE tokeniser with vocabulary size \(V = 4096\) on the same corpus. Documents are concatenated with EOS separators and chunked into fixed-length windows of length \(n_{\mathrm{ctx}} = 512\).

\subsubsection{Architecture}

We use a decoder-only, attention-only language model with bilinear attention and no MLP blocks. The model has \(L = 2\) layers, \(H = 8\) attention heads, and model dimension \(d_{\mathrm{model}} = 256\). The embed and unembed are bias-free and not tied. Positional information is added using RoPE inside the attention layers.

To keep the model structure bilinear, we use only a scalar normalisation method applied before the unembedding. Specifically, we normalise the sequence by a scalar \(s\) computed from the first token's hidden state. 
\[
s = \sqrt{\frac{1}{d_{\mathrm{model}}}\sum_{j=1}^{d_{\mathrm{model}}} h_{0,j}^2 + \epsilon}.
\]

\begin{table}[h]
\centering
\begin{tabular}{l r r}
\hline
\textbf{Weight} & \textbf{Shape} & \textbf{Parameters} \\
\hline
\texttt{embed.weight} & \((V, d_{\mathrm{model}}) = (4096, 256)\) & \(1{,}048{,}576\) \\
\hline
\texttt{attn.q1.weight} & \((256, 256)\) & \(65{,}536\) \\
\texttt{attn.q1.bias}   & \((256,)\)     & \(256\) \\
\texttt{attn.k1.weight} & \((256, 256)\) & \(65{,}536\) \\
\texttt{attn.k1.bias}   & \((256,)\)     & \(256\) \\
\texttt{attn.q2.weight} & \((256, 256)\) & \(65{,}536\) \\
\texttt{attn.q2.bias}   & \((256,)\)     & \(256\) \\
\texttt{attn.k2.weight} & \((256, 256)\) & \(65{,}536\) \\
\texttt{attn.k2.bias}   & \((256,)\)     & \(256\) \\
\texttt{attn.v.weight}  & \((256, 256)\) & \(65{,}536\) \\
\texttt{attn.o.weight}  & \((256, 256)\) & \(65{,}536\) \\
\hline
Layer subtotal & -- & \(394{,}240\) \\
All layers \((\times L = 2)\) & -- & \(788{,}480\) \\
\hline
\texttt{final\_norm} \((\texttt{tok0\_batch})\) & -- & \(0\) \\
\texttt{unembed.weight} & \((V, d_{\mathrm{model}}) = (4096, 256)\) & \(1{,}048{,}576\) \\
\hline
\textbf{Total} & -- & \(\mathbf{2{,}885{,}632}\) \\
\hline
\end{tabular}
\caption{Attention-only language model parameter count}
\end{table}

\subsubsection{Loss}

For a token sequence \(S_i = (t^i_1,\ldots,t^i_T)\), the empirical next-token loss at position k is
\[
\ell_{n,k}(w)
=
\frac{1}{n}
\sum_{i=1}^{n}
-\log \mathrm{softmax}\!\left(f_w(S^i_{\leq k})\right)\!\left[t^i_{k+1}\right],
\qquad
k = 1,\ldots,T-1.
\]
The training loss is
\[
\ell_n(w)
=
\frac{1}{T-1}
\sum_{k=1}^{T-1}
\ell_{n,k}(w),
\qquad
T = n_{\mathrm{ctx}} = 512,
\]
with held-out loss defined analogously on the validation split.

\subsubsection{Training}\label{app:LanguageTrainingDetails}

Training is performed as a single streaming pass over the shuffled, EOS-separated Pile stream. The main run uses \(T_{\mathrm{train}} = 20{,}000\) Optimiser steps, batch size \(B = 384\), and context length \(n_{\mathrm{ctx}} = 512\), corresponding to approximately \( 4 * 10^9\) training tokens. Evaluation is performed using a cached validation split from the Pile.

Optimisation uses Muon for the internal attention projection weights, while the attention output projection, embedding, and unembedding parameters are optimised with AdamW. The Muon learning rate is \(0.02\), and the AdamW learning rate is \(3 \times 10^{-4}\), with Adam betas \((0.9, 0.95)\), weight decay \(0.1\), and gradient clipping at \(1.0\). Both optimiser groups use a cosine schedule with a 500-step warmup and decay to \(0.2\) times the peak learning rate by step \(20{,}000\). The model took 5 hours to train on A10.

\begin{table}[htbp]
\centering
\begin{tabular}{lll}
\hline
\textbf{Hyperparameter} & \textbf{Category} & \textbf{Value} \\
\hline
Dataset & Data & DSIR-filtered Pile-50M \\
\(V\) & Data & \(4096\) \\
\(n_{\mathrm{ctx}}\) & Data & \(512\) \\
\(B\) & Data & \(384\) \\
\(N_{\mathrm{steps}}\) & Data & \(20{,}000\) \\
Tokens seen & Data & \(\approx 3.93 \times 10^9\) \\
\(L\) & Model & \(2\) \\
\(H\) & Model & \(8\) \\
\(d_{\mathrm{model}}\) & Model & \(256\) \\
Attention type & Model & bilinear \\
Attention scale & Model & \(0.35\) \\
RoPE base & Model & \(10{,}000\) \\
Norm & Model & \texttt{tok0\_batch} before unembedding \\
Optimiser & Optimisation & Muon + AdamW \\
Muon LR & Optimisation & \(0.02\) \\
AdamW LR & Optimisation & \(3 \times 10^{-4}\) \\
Weight decay & Optimisation & \(0.1\) \\
Scheduler & Optimisation & cosine decay with 500-step warmup \\
\hline
\end{tabular}
\caption{Language-modeling transformer training, data, model, and optimisation hyperparameters.}
\label{tab:language_model_hyperparameters}
\end{table}
\subsection{Metrics tracked over training}

\subsubsection{N-gram metrics}

The n-gram metric measures whether the model has learned common fixed-length token patterns from the training data. First, the most frequent n-grams \((t_1,\ldots,t_n)\) are extracted from the training corpus. For each n-gram, the model is given the prefix \((t_1,\ldots,t_{n-1})\) and evaluated on its ability to predict the final token \(t_n\). Averaging this cross-entropy over frequent training n-grams gives the n-gram memorisation loss, denoted \(l_{\mathrm{ngram}}(n)\). The same prediction task is then evaluated on validation sequences, giving \(l_{\mathrm{test}}(n)\). The final score is
\[
\mathrm{ngram\ score}(n) = \frac{l_{\mathrm{test}}(n)}{l_{\mathrm{ngram}}(n)}.
\]
A lower ratio indicates that the model performs similarly on validation n-gram prediction and frequent training n-gram prediction, while a higher ratio suggests stronger specialization to the frequent training patterns.

\section{Robustness to non-Gaussian input distributions} \label{app:construct-validity}

In \autoref{sec:gaussian_metric}, we showed that tensor similarity coincides with a canonical behavioural metric, cosine similarity of empirical outputs, under the assumption of a perfectly Gaussian input distribution. But how robust is this to non-Gaussian input distributions? In this appendix, we investigate the question statistically.

\paragraph{Dataset} We consider a toy task where models are trained to detect the index of the second-highest value (or \textit{2nd argmax}) in a list of four numbers sampled from one of the following 9 input distributions:
\begin{itemize}
    \item Gaussian: Standard Gaussian distribution (mean 0 and variance 1). 
    \item Half Gaussian: Absolute value after sampling from a standard Gaussian distribution.
    \item Bimodal: Combination of two Gaussians designed to be bimodal.
    \item Uniform: Uniform distribution on $[-1, 1]$.
    \item Laplace: Laplace distribution with location parameter $\mu = 0$ and scale parameter $b = \frac{1}{\sqrt{2}}$. 
    \item `Sparse Spikes': A distribution where there is a 75\% chance of drawing 0 and a 25\% chance of drawing from a Gaussian distribution with mean 1 and variance 4.
    \item Permutations: The set of permutations of the list (1, 2, 3, 4).
    \item Correlated Gaussian: Standard Gaussian distribution with a correlation introduced between the four inputs, defined by a fixed positive-definite correlation matrix.
    \item `Gaussian and $-10$': Standard Gaussian distribution for the first three inputs; a constant $-10$ for the last input.
\end{itemize}

\textit{Example:} One possible sample from the `Gaussian and $-10$' distribution is the following: $[-0.54, -0.19, 0.17, -10.00]$. For this data point, the correct output for the model is $1$, since the second-highest value is $-0.19$, which lies at index $1$.

\paragraph{Model Architecture} The model consists of a toy one-layer bilinear network. No residuals, normalisation layers, or biases are included in the model.

\paragraph{Model Dimensions and Training} \autoref{tab:dist-robustness-params} summarises the model dimensions and training hyperparameters. Across all five seeds, models trained on the three distributions which are \textit{not} symmetric about zero (namely Half Gaussian, Permutations, and `Gaussian and $-10$') achieved accuracy above $90\%$, while models trained on the six symmetric distributions plateaued at approximately $50\%$ (with the exception of the `Sparse Spikes' models, which plateaued around $39\%$). This is due to a fundamental limitation in the pure, single-layer bilinear architecture used: it cannot distinguish $x$ from $-x$, preventing the model from meaningfully exceeding $50\%$ accuracy on any distribution symmetric about zero.

  \begin{table}[tbh!]
      \centering                                                                
      \begin{tabular}{l@{\hspace{2cm}}r}                    
          \toprule
          \multicolumn{2}{c}{\textbf{Architecture}} \\
          \midrule                                                              
          \addlinespace[2pt]
          \textbf{input dim} $n$   & 4  \\                                      
          \textbf{rank}            & 32 \\                  
          \textbf{n\_layer}        & 1  \\                                      
          \addlinespace[2pt]
          \midrule                                                              
          \multicolumn{2}{c}{\textbf{Training Parameters}} \\
          \midrule                                                              
          \addlinespace[2pt]
          \textbf{dropout}         & 0.0       \\                               
          \textbf{weight decay}    & 0.0       \\           
          \textbf{batch size}      & 512       \\                               
          \textbf{learning rate}   & $10^{-2}$ \\
          \textbf{optimizer}       & Adam      \\                               
          \textbf{schedule}        & constant  \\                               
          \textbf{train steps}     & 4096      \\
          \textbf{seeds}           & 5         \\                               
          \addlinespace[2pt]                                                    
          \bottomrule
      \end{tabular}                                                             
      \caption{Model architecture and training setup for the distribution
  robustness experiment.}                                                       
      \label{tab:dist-robustness-params}
  \end{table}

\begin{figure}[tbh!]
    \centering
     \includegraphics[width=0.9\linewidth]{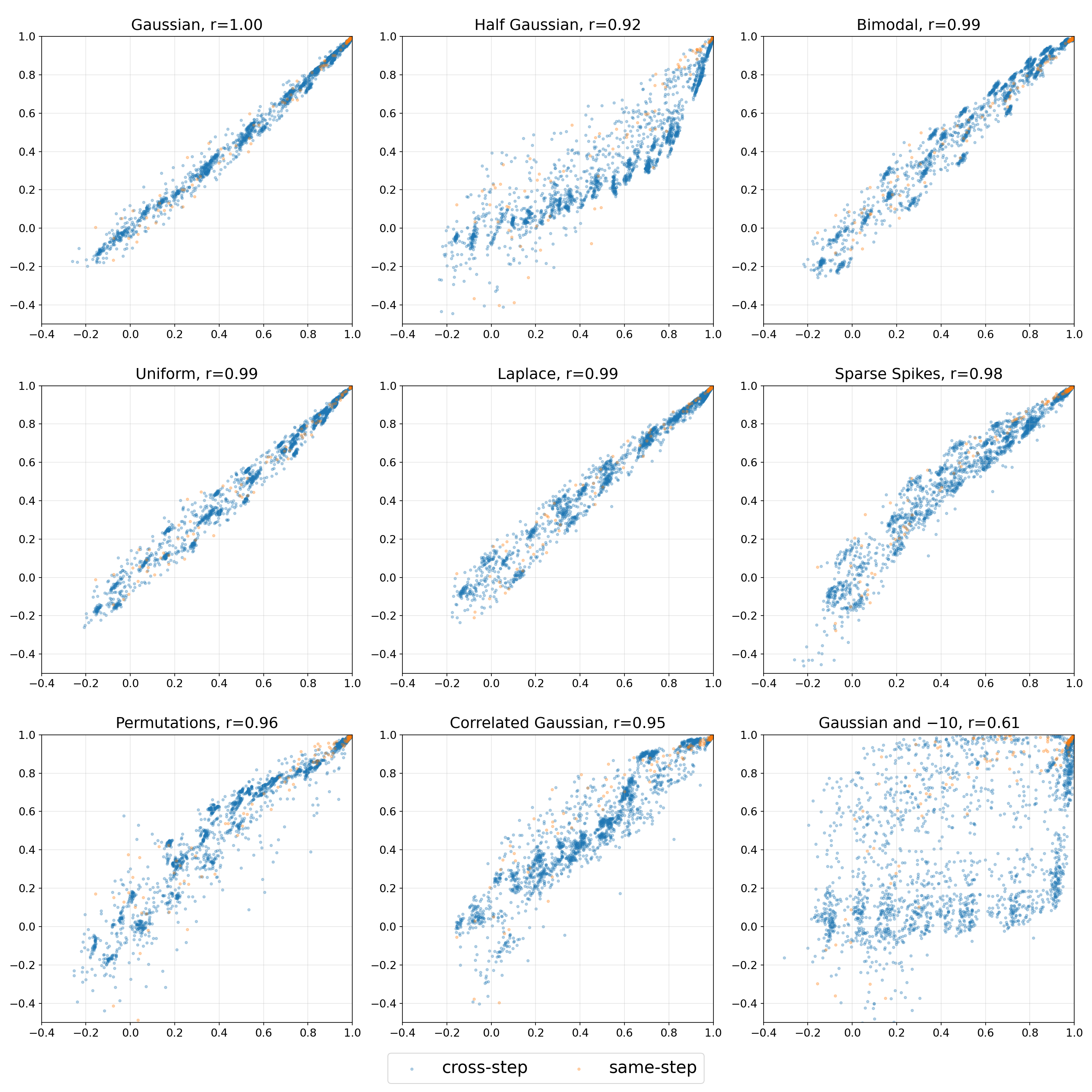}
    \caption{Five one-layer bilinear models were trained for each of nine input data distributions, and they were each sampled at 14 checkpoints. Pairs of these models were compared with a tensor-based similarity metric (shown on the x-axis) and a behavioural, output-based cosine simila rity metric (shown on the y-axis). Pearson correlations are displayed above each scatterplot.}
    \label{fig:scatter-grid-distributions}
\end{figure}

\autoref{fig:scatter-grid-distributions} shows Pearson correlations between tensor similarity and the cosine similarity of empirical outputs across different training checkpoints and seeds. While the correlation between tensor similarity and output similarity is strongest for Gaussian inputs, it exceeds 0.9 for eight out of the nine distributions tested, including distributions of varying shape and one with statistical dependence between input dimensions.

The exception is the correlation across the `Gaussian and $-10$' dataset, with a measured correlation of $r$ = 0.61. Intuitively, the reason for this discrepancy is that here, the last index is so low that it will virtually never point to the second highest value, so the model learns to never output that index. This means that a substantial portion of network weights do not influence the ultimate result, and the weight-based similarity metric we use here is less correlated with the empirical similarity.

These results suggest that the tensor similarity metric is a robust proxy for empirical similarity across a broad range of distributions, not just Gaussians, so long as the weights all meaningfully contribute toward the output.

\end{document}